# Model-based Transfer Learning for Automatic Optical Inspection based on domain discrepancy

Erik I. Valle Salgado[a], Haoxin Yan[a], Yue Hong[a], Peiyuan Zhu[c], Shidong Zhu[c], Chengwei Liao[d], Yanxiang Wen[e], Xiu Li[b], Xiang Qian[b], Xiaohao Wang[b], Xinghui Li [a]*.
[a]Tsinghua-Berkeley Shenzhen Institute, University Town of Shenzhen, Nanshan District, Shenzhen 518055, P.R.China;
[b]Tsinghua Univ. Shenzhen Int. Grad. School, University Town of Shenzhen, Nanshan District, Shenzhen 518055, P.R.China;
[c]Shenzhen Han Industrial Technologies Co., Ltd., Shenzhen, P.R. China;
[d]Guilin Han Smart Instrument Co., Ltd., Qixing District, Guilin, P.R. China;
[e]Guangxi Guihua Intelligent Manufacturing Co., Ltd., Nanning 530022, P.R. China

## ABSTRACT

Transfer learning is a promising method for AOI applications since it can significantly shorten sample collection time and improve efficiency in today's smart manufacturing. However, related research enhanced the network models by applying TL without considering the domain similarity among datasets, the data long-tailedness of a source dataset, and mainly used linear transformations to mitigate the lack of samples. This research applies model-based TL via domain similarity to improve the overall performance and data augmentation in both target and source domains to enrich the data quality and reduce the imbalance. Given a group of source datasets from similar industrial processes, we define which group is the most related to the target through the domain discrepancy score and the number of samples each has. Then, we transfer the chosen pre-trained backbone weights to train and fine-tune the target network. Our research suggests increases in the F1 score and the PR curve up to 20% compared with TL using benchmark datasets.

## 1. INTRODUCTION

In Automatic Optical Inspection, defect detection and classification in products from multiple areas is critical in ensuring the quality of industrial products. Due to the incremental popularity and development of computer vision methods, researchers try implementing the latest technologies related to defect detection and feature extraction algorithms. Abd Al Rahman M. Abu Ebayyeh and Alireza Mousavi [1] reviewed research articles that conducted AOI systems and algorithms to detect defects in commonly inspected components in the electronics industry during the last two decades. It covers multiple defect features, image acquisition techniques, inspection algorithms, and sorting mechanisms. They highlighted various methods for object recognition, but we will focus on Convolutional Neural Networks since they outperform prolonged and repetitive activities compared with RNNs [2], offer options beyond detection and classification methods, extract features to improve the generalization, are computationally efficient and suitable for parameter tuning.

In recent years, multiple research papers have paid attention to enhancing elements of neural network architectures to meet the criteria required to detect minor and varied defects proper of Automatic Optical Inspection. For instance, Rui Huang et al. [3] modified the YOLOv3 model for detecting electronic components in complex backgrounds by adding a confidence score on each bounding box through logistic regression, replacing the loss function with BCE and independent logistic, and substituting the Darknet53 backbone with Mobilenet classifiers. Yibin Huang et al. [4] proposed a model for saliency detection of surface defects that consists of MCue to generate resized inputs, U-Net, and Push networks to define the specific location of predicted defects. Based on YOLOv4 architectures and their applications in defect detection and classification of rail surfaces, Noreen Anwar et al. [5] changed the activation functions of the CSPDarknet-53 with SELU, used SAM at upsampling and downsampling points, and redefined the loss function in terms of object classification, object confidence, and object location offset, all with balance coefficients. Junjie Xing and Mingping Jia [6] created a CNN backbone for the classification model (SCN) with symmetric modules plus three convolution branches with an FPN structure for feature identification, adding an optimized IoU as the loss function. Clearly, the previous publications show how authors modify models without considering real-world data quality, distribution, and similarity among samples [7] [8]. Usually, defect inspection tasks involve imbalanced datasets with a long-tailed and open-ended distribution. A classifier must categorize among majority and minority classes, generalize from a few known instances and recognize novelty upon a never observed input [9]. Other research in this field pointed out typical techniques to deal with the lack of samples by data augmentation [10], [11], transfer learning-aided models pretrained on benchmark datasets without domain similarities with the target domain [12], [13], specialized network structures to extract meaningful features from the targets [14], or fine-tuning the transferred weights [15].

*li.xinghui@sz.tsinghua.edu.cn

The main contributions of our research are summarized as follows:

- In the context, we explore if and how the existing defect-detection datasets and benchmarks can be applied explicitly in model-based transfer learning for Automatic Optical Inspection. Since metal surface defect detection and other optical inspection datasets differ from standards in colorspace, categories, and data distribution, we aim to enhance the quality of pre-trained weights to achieve better results.

- We propose the domain discrepancy score and source domain selection method based on the Wasserstein distance, the Gini distance, and class overlapping to measure the difference among multiple distributions and obtain a score that describes the similarity among datasets. The lower the score, the closer the resemblance between the datasets. This algorithm lets us choose a similar source dataset for a target domain together with data augmentation techniques to match the source dataset with the target as closely as possible.

## 2. METHODOLOGY

Transfer learning aims to enhance the performance of the hypothesis function for a target task by discovering and transferring latent knowledge from a source domain and domain tasks, where neither the domains nor the source and target duties must be equivalent. Thus, transfer learning relaxes the hypothesis that such training data must be independent and identically distributed with the test data. However, real-world data usually contain structures among the data instances. and samples in categories are typically insufficient. Hence, transfer learning is suitable to overcome the reliance on large quantities of high-quality data by leveraging helpful information from other related domains.

Previous works in this field have demonstrated the effectiveness of transferring pretrained weights using benchmark datasets such as COCO and ImageNet. Indeed, they all showed relevant improvements in precision, recall, and other derived metrics and reductions in computing time and resources. Nevertheless, their source datasets may not be related to the target samples in terms of colorspace, categories, image size, or even skewed class instances. Thus, our proposal rates the source domain's similarity with respect to the target through a domain discrepancy score. It evaluates the domain discrepancy from the available source dataset with the target domain so that we can select the best match and so adapt the source domain appropriately.

Based on the binning method proposed by Tianyu Han et al. [16], our algorithm keeps categories from the source dataset that overlap with the target samples and removes the rest. We use this brand class-similar source dataset to train a model whose earlier layers (backbone) contain more generic features suitable to fine-tune the latter model that comprises the target domain. Since the target dataset and any other source domain may have severe data imbalance, data augmentation over minorities increases the number of samples and partially mitigates this issue. Likewise, instances from the majority classes could be redundant or cause noise due to their resemblance with the target. Adjusting their quantity through undersampling relieves data skewness and leaves enough samples that match the target dataset.

### 2.1 Earth Mover's Distance as a metric for domain similarity

Since a domain $D^{(i)} = \{\mathbb{X}^{(i)}, P(\mathbb{X}^{(i)})\}$ comprises a feature space $\mathbb{X}^{(i)}$ and a marginal probability distribution $P(\mathbb{X}^{(i)})$, and a task $T^{(i)} = \{Y^{(i)}, f(\cdot)\}$ also contains a label space $Y^{(i)}$ and conditional probability distribution $f(\cdot)$, research on this topic usually assumes that domains share the same feature space or simply do not consider it. On the other hand, if the domains have distinct feature spaces or label spaces, one has to project the data onto the same feature or label space and then use the statistical distance estimations as a follow-up step [17]. This paper aims to match similarities among distribution and reduce its gap through a two-step domain screening based on the EMD distance and a modified Bin Similarity algorithm based on [16].

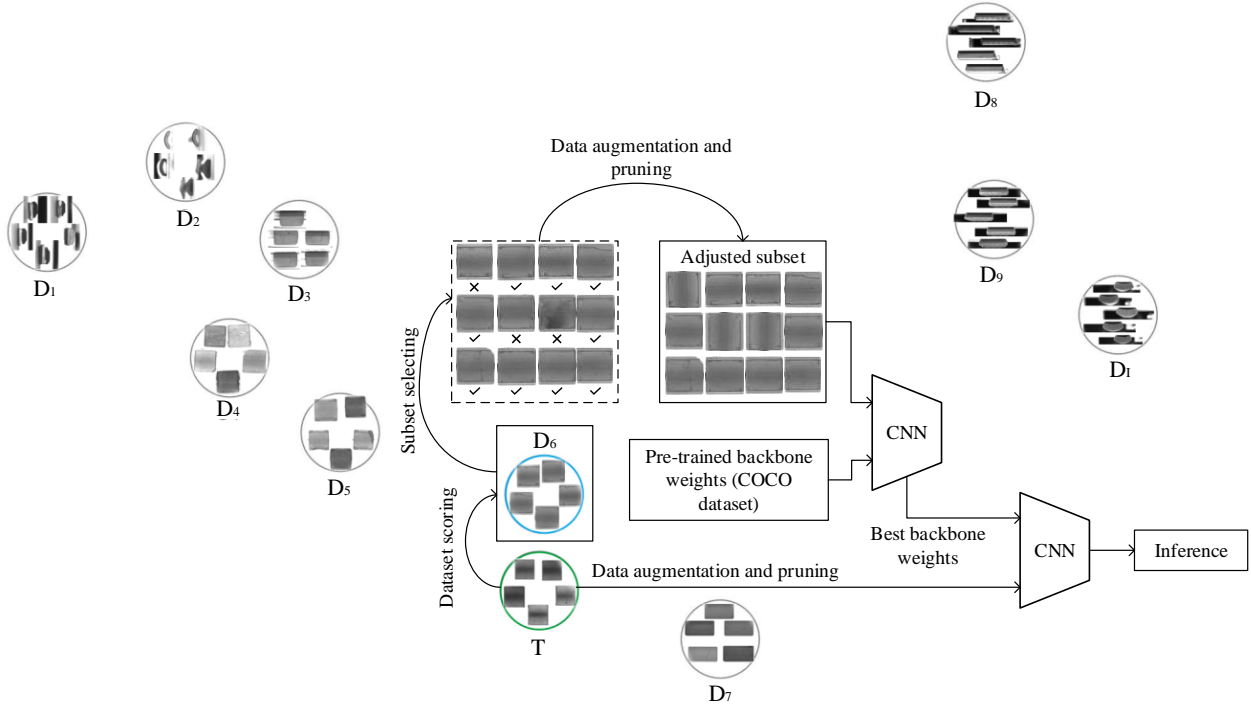

Figure 1 Overview of our domain similarity and subset selection scheme. The system consisted of five parts. (1) The source dataset scoring executed the domain discrepancy analysis to select a source domain dataset from the repository close to the target dataset regarding EMD distance, categories overlapping, the number of samples, and the Gini score. (2) The dataset subset selection algorithm minimized the divergence between domains by removing samples from the source dataset to match as many categories as possible. (3) Data augmentation mitigated the data imbalance among categories by increasing minority classes' samples. (4) Pretraining the network on the selected sub-dataset using benchmark pre-trained weights. (5) Transference of the best pre-trained weights that belong to the network backbone to the target network as initialized the rest randomly. Then, trained the last network using a data-augmented target dataset.

## 2.2 Source dataset scoring and selection

**Signatures estimation**

Each dataset image $x_m^{(i)}$ possesses many pixels that characterize a distribution and describe the artifacts contained in a picture. Representing the overall distribution of the features of all the photos in the database with bins, either fixed (histogram) or adaptive, may not attain an equilibrium between expressiveness and efficiency [18]. Instead, using variable-size descriptions of distributions or signatures provides an alternative that takes advantage of the dominant clusters extracted from the original distribution to form a compressed representation. A signature comprises the main clusters of a distribution with a weight that denotes the size of each cluster. A clustering approach like the k-means objective helps us set such pixels of an image $x_m^{(i)}$ in $K$ clusters and then group a full dataset $X_i = \{x_1^{(i)}, \ldots, x_M^{(i)}\}$ into a few cohesive groups.

Although our samples include label spaces, we temporarily omit them in the first step for all datasets. We want to represent them as a whole instead of selecting the most similar classes from the source domain to the target domain, as in [16]. The reason is that given a target dataset, we must pick one of the available magnetic tile datasets, which usually share a relevant number of classes because their fabrication process, environment, and material properties are similar.

Mathematically, given input image datasets $X_i$, $X_T$, and the number of bins $K$, the signatures estimation algorithm can be defined as a function $SE(X_i, K)$, which computes a signature for each dataset plus their respective weights. Algorithm 1 presents the SE function.

Table 1. Signatures Estimation Algorithm for domain and target datasets.

| **Algorithm 1** Signatures Estimation |
|---|
| **Input**: |
| $X_1, \ldots, X_I, X_T$: $I$ sources and a target datasets. $i \in \{1, \ldots, I\}$ |
| $x_1^{(i)}, \ldots, x_M^{(i)}$: M input images for dataset $i$. $m \in \{1, \ldots, M\}$ |
| **Output**: |
| $\left\{ s^{(i)} = \left( \left( c_1^{(i)}, \ldots, c_K^{(i)} \right), \left( w'^{(i)}_1, \ldots, w'^{(i)}_K \right) \right) \right\}$ main $K$ centroids and weights. |
| **Statement**: |
| For each dataset $X_i, X_T$: |
|     1. Define the average number of centroids $K$ per dataset to generate through the distortion function $RSS$. |
|     2. Calculate centroids $c_{m,k}^{(i)}$ and weights $w_{m,k}^{(i)}$ for all images $x_m^{(i)}$ using k-means. |
|     3. Average k-means clusters $c_{m,k}^{(i)}$ and assign normalized weights $w_k^{(i)}$ to each mean cluster $\omega_k^{(i)}$. |

We first define the number of clusters to form and the number of centroids to generate. For this purpose, the k-means objective (distortion function) minimizes samples' average squared Euclidean distance from their cluster center. Such a center is the mean centroid $c_{m,k}^{(i)}$ of the instances in a cluster $\omega_{m,k}^{(i)}$:

$$c_m^{(i)} = kmeans(x_m^{(i)}) = \left\{ c_{m,1}^{(i)}, c_{m,2}^{(i)}, \ldots, c_{m,K}^{(i)} \right\} \quad (1)$$

$$c_{m,k}^{(i)} = \frac{1}{|\omega_{m,k}^{(i)}|} \sum_{\vec{x}_m^{(i)} \in \omega_{m,k}^{(i)}} \vec{x}_m^{(i)} \quad (2)$$

Where $\vec{x}_m^{(i)}$ is a pixel value in the RGB color space belonging to picture $x_m^{(i)}$, each image contains $weight \times height$ pixels $\vec{x}_m^{(i)}$. Our proposal takes advantage of the triangle inequality to accelerate k-means [19] since it becomes more effective as the number K of clusters increases, which opens the possibility of thorough similarity analysis. We can take any optimized k-means algorithm to calculate the proper quantity of K centroids through the residual sum of squares (RSS). This function measures how well the centroids denote the members of their clusters via the squared distance of each $\vec{x}_m^{(i)}$ from its mean centroid summed over all $\vec{x}_m^{(i)}$ belonging to cluster $\omega_{m,k}^{(i)}$:

$$RSS_k = \sum_{\vec{x}_m^{(i)} \in \omega_{m,k}^{(i)}} \left| \vec{x}_m^{(i)} - c_{m,k}^{(i)} \right|^2 \quad (3)$$

$$RSS = \sum_{k=1}^{K} RSS_k \quad (4)$$

Since each dataset has thousands of images, calculating K with a small sampling reduces the computing time and resources thanks to the similarity among elements of the same domain. Indeed, this premise applies only if the pictures belong to the same product category and share similar capturing conditions. Otherwise, image pre-processing may mitigate contrast, illumination, and distortion variations. Moreover, if the domain consists of different products, we can search for similar sub-datasets from the source domain and split it accordingly.

Having specified a fixed number of K centroids, we calculate the K-means for all images of datasets as in the previous steps, excluding the RSS estimation. Once the final $c_{m,k}^{(i)}$ for each picture is obtained, we can define the weight of cluster $w_{m,k}^{(i)}$ as the ratio of pixels $\vec{x}_m^{(i)}$ that belong to it, being the total sum of such parameters 1.

$$\sum_{k=1}^{K} w_{m,k}^{(i)} = 1 \tag{5}$$

$$w_m^{(i)} = \left\{ w_{m,1}^{(i)}, w_{m,2}^{(i)}, \ldots, w_{m,K}^{(i)} \right\} \tag{6}$$

Thus, a signature comprises both centroids and weights to form a compressed representation of an image.

$$\left\{ s_m^{(i)} = \left( c_m^{(i)}, w_m^{(i)} \right) \right\} \tag{7}$$

Since each image has its signature and they all form a dataset, we propose to denote the $M$ signatures in only $K$ main clusters $c_k^{(i)}$ using k-means but considering all $c_m^{(i)}$ and their weights $w_m^{(i)}$ as inputs instead of pixel values and frequency. The main weights $w_k^{(i)}$ may not meet the constraint in equation 5, so adjusting their values helps meet the prior criterion.

$$w'^{(i)}_k = \frac{w_k^{(i)}}{\sum_k w_k^{(i)}} \tag{8}$$

Finally, the signature that represents a dataset is described as follows:

$$\left\{ s^{(i)} = \left( \left( c_1^{(i)}, \ldots, c_K^{(i)} \right), \left( w'^{(i)}_1, \ldots, w'^{(i)}_K \right) \right) \right\} \tag{9}$$

**Earth Mover's Distance for domain discrepancy estimation among datasets**

The Earth Mover's Distance measures the difference between the source and target domains with features. Our approach provides a ground distance that assesses dissimilarity between datasets signatures to address the problem of lifting these metrics from individual elements to distributions. It implies obtaining distances between picture color distributions in colorspace terms. Indeed, the solution is the minimum amount of "work" required to transform one signature into the other [18]. We also aim to allow these distances for partial matches to compare one distribution with a subset of another.

Formally, the EMD is a linear programming problem: Let $s^{(i)} = \left\{ \left( c_1^{(i)}, w_1^{(i)} \right), \ldots, \left( c_M^{(i)}, w_M^{(i)} \right) \right\}$ be the first signature with $M$ clusters; $s^{(j)} = \left\{ \left( c_1^{(j)}, w_1^{(j)} \right), \ldots, \left( c_N^{(j)}, w_N^{(j)} \right) \right\}$ the second signature with $N$ clusters; and $G = [g_{u,v}]$ the ground distance matrix where $g_{u,v}$ is the ground distance between clusters $c_m^{(i)}$ and $c_n^{(j)}$. This last measure is simply the Euclidean distance in the color space $g_{u,v} = \sqrt{(R_u - R_v)^2 + (G_u - G_v)^2 + (B_u - B_v)^2}$ between the analyzed clusters. Next, we must find a flow $F = [f_{u,v}]$, with $f_{u,v}$ the flow between $c_m^{(i)}$ and $c_n^{(j)}$, that minimizes the overall cost and distance.

$$WORK(X_i, X_T, F) = \sum_{u=1}^{M} \sum_{v=1}^{N} g_{u,v} f_{u,v} \tag{10}$$

Subject to the following constraints

$$f_{u,v} \geq 0 \quad 1 \leq u \leq M, \quad 1 \leq v \leq N \tag{11}$$

$$\sum_{v=1}^{N} f_{u,v} \leq w'^{(i)}_u \quad 1 \leq u \leq M \tag{12}$$

$$\sum_{u=1}^{M} f_{u,v} \leq w'^{(j)}_v \quad 1 \leq v \leq N \tag{13}$$

$$\sum_{u=1}^{M} \sum_{v=1}^{N} f_{u,v} = \min \left( \sum_{u=1}^{M} w'^{(i)}_u, \sum_{v=1}^{N} w'^{(j)}_v \right) \tag{14}$$

So far, the EMD calculation demands an initial flow $F_0$ close enough to the final solution so that we save computational time and resources. To do so, we took the work of Edward Russell [20], which extended Dantzig's algorithm (simplex) to calculate a starting basis for the transportation problem that produces a near-optimal basis, and then we optimize it through the Sequential Least-Squares Programming (SLSQP) [21]. Finally, we can obtain the Earth Mover's Distance with the optimal flow $F$ previously estimated as the subsequent work normalized by the total flow:

$$EMD(s_i, s_T) = \frac{\sum_{u=1}^{M} \sum_{v=1}^{N} g_{u,v} f_{u,v}}{\sum_{u=1}^{M} \sum_{v=1}^{N} f_{u,v}} \tag{15}$$

**Long-tailedness Metrics: The Gini Coefficient**

Measuring the long-tailedness of data is a relevant issue to solving the long-tailed visual recognition problem. Although there are multiple long-tailedness metrics, such as the imbalance factor, standard deviation, or mean/median, the Gini coefficient [22] can effectively differentiate long-tailed and balanced datasets. The reason is that it is not affected by extreme samples, the absolute number of data, and has a bounded distribution (0,1) [23].

To obtain it, we must follow the next steps:

1. Compute the normalized cumulative distribution $\{\mathbb{C}_i\}$, assuming that $\mathbb{k}$ categories and their respective number samples $m_i, (i = 1, 2, \ldots, \mathbb{k})$ are in ascending order:

$$\mathbb{C}_i = \frac{1}{\mathbb{k}} \sum_{j=1}^{i} m_j \tag{16}$$

2. Calculate the area $B$ under the Lorenz Curve $L(x), x \in [0,1]$:

$$L(x) = \begin{cases} \mathbb{C}_i, & x = \frac{i}{\mathbb{k}} \\ \mathbb{C}_i + (\mathbb{C}_{i+1} - \mathbb{C}_i)(\mathbb{k}x - i), & \frac{i}{\mathbb{k}} < x < \frac{i+1}{\mathbb{k}} \end{cases} \tag{17}$$

$$B = \int_0^1 L(x) dx = \sum_{i=1}^{\mathbb{k}} \frac{\mathbb{C}_i + \mathbb{C}_{i-1}}{2} \cdot \frac{1}{\mathbb{k}} \tag{18}$$

$$A = 0.5 - B \tag{19}$$

3. Estimate the Gini coefficient:

$$\delta = \frac{A}{A + B} > 0 \tag{20}$$

**Domain discrepancy score and dataset selection**

Now that we have a premise to describe the similarity across datasets, we must ponder a function that describes the similarity between the target dataset and the available sources. Thus, we propose a function that consists of the product of the EMD and the Gini score divided by the set cardinality of the label space intersection of the target and the source tasks. Since a dataset for object detection and classification has multiple instances belonging to the same category and its cardinality is proportional to the sum of samples per category, we convert such multiset $Y^{(i)}$ to a set $Y'^{(i)}$ by simply taking the intersection of our multisets with the universe $U$, which outcomes the number of categories instead of elements, equilibrating the relevance of all categories in a dataset. Finally, the lower the score, the more similar the datasets are.

$$disc(D^{(T)}, D^{(i)}) = \frac{EMD(s_i, s_T) \cdot \delta}{|Y'^{(T)} \cap Y'^{(i)}| + \varepsilon} \tag{20}$$

Finally, we pick the three most minor scores and choose the one with more samples. The reason is that more images and labels result in higher performance in terms of the evaluation metrics described in section 3.2.

### 2.4 Target Label-Space Conditioned Subset selection

In traditional machine learning, learning a model given a set of training samples to find an objective function assumes that both training and test data come from the same distribution and share a similar joint probability distribution. On the other hand, conventional domain adaptation aims to solve the prediction function $f_T(\cdot)$ of the target task $T_T$ in the target domain $D_T$ through the knowledge acquired by the source domain $D_i$ and the source task $T_i$ if the feature and label spaces remain unchanged as their probability distributions may change between domains [24]. In manufacturing applications and other real-world matters violating this last constraint is feasible since the datasets can emerge from different tasks or distributions. Hence, easing the label space gap by removing samples containing tags outside the target task is the first step to reducing such discrepancy. Our purpose is to make the data in the source particular categories a subset of the target label set. This idea is worthy given a group of datasets that could be considered subsets of the target label set and thus enhance the target model performance by taking all the source domain knowledge included in the shared classes. Some authors, such as Saito et al. [25] and Hong Liu et al. [26], took advantage of the previous premise and utilized either adversarial training or binary classifiers to align target samples with source-known elements or reject them as unknown target ones. Since our datasets contain tags with the same names, we are not considering further processing and removing the source samples that did not match the target task space and kept the rest as they were.

### 2.5 Data Augmentation

Generalization is part of any deep convolutional networks since their objective is to improve the performance of a network trained on previously seen data versus never seen samples. Indeed, models with poor generalization tend to overfit the training data, so augmenting the number of samples represents a way to expand a dataset, save labeling costs, and improve classification performance. Although Data Augmentation cannot overcome all biases present in the minority classes, this technique prevents or significantly lessens multiple biases such as occlusions, lighting, scales, or changes in the background. Our scope covers basic image manipulations, including flipping, rotation, contrast, and noise injection, because those changes are typical of a manufacturing environment. Nevertheless, these methods only create data by image-level linear transformations and may not represent new distributions introduced by unknown defects with changes in the defects' shape or lighting orientations [27].

### 2.6 Transfer Learning

Transfer learning aims to enhance the performance of the hypothesis function for a target task by discovering and transferring latent knowledge from a source domain and domain tasks, where neither the domains nor the source and target duties must be equivalent. Hence, this technique is suitable to overcome the reliance on large quantities of high-quality data by leveraging helpful information from other related domains. Kim et al. [15] explored a couple of weight transference methods from a source network using ImageNet 2012 to a target network with data provided by DAGM [28] either freezing them or just fine-tuning them. Our proposal not only fine-tunes a network by using a benchmark dataset but also uses a close domain in terms of the domain discrepancy score. As shown in Figure 1, we first trained the closest dataset with pre-trained weights on COCO by transferring only those weights belonging to the network backbone and initialize the rest randomly. Again, we transfer these domain network backbone weights to the target network and started the remaining layer weights randomly. Finally, fine-tuning the transferred weights was a key point to obtain the highest performance compared with other methods, as stated in [15] .

## 3. EXPERIMENTS

### 3.1 Datasets

It is a group of real-world class-imbalanced datasets containing 47561 gray-scale images with 20 different defect categories manually labeled arranged in 11 magnetic tile datasets, as shown in Table 1. They all follow a typical long-tailed distribution, where the one with the highest imbalanced rate used in this paper corresponds to jy-381-2, and the lowest belongs to dc-1. Pictures do not have any preprocessing, so the dataset can reflect the distribution of multiple defect types on a production line.

Table 1. Magnetic tile datasets: number of samples per dataset and category.

| Label | Defects | Dataset | | | | | | | | | | | Sum |
|---|---|---|---|---|---|---|---|---|---|---|---|---|---|
| | | dc-1 | jy-381-2 | jy-381-4 | lc-101 | lc-201 | nj-101 | nj-201 | xh-1 | xh-2 | xh-3 | xh-4 | |
| $y_0$ | white crack | 0 | 0 | 0 | 0 | 0 | 0 | 0 | 0 | 267 | 0 | 0 | 267 |
| $y_1$ | standard chipping | 0 | 3041 | 1676 | 3875 | 4070 | 1021 | 1944 | 3102 | 1515 | 1209 | 1613 | 23066 |
| $y_2$ | standard crack | 349 | 62 | 250 | 2178 | 631 | 3058 | 2787 | 1403 | 1002 | 341 | 54 | 12115 |
| $y_3$ | chamfer | 0 | 1 | 212 | 84 | 393 | 1 | 2 | 40 | 0 | 0 | 0 | 733 |
| $y_4$ | multifaceted | 2 | 3 | 0 | 2778 | 294 | 705 | 762 | 373 | 138 | 272 | 0 | 5327 |
| $y_5$ | crystallization | 0 | 0 | 0 | 0 | 0 | 0 | 0 | 14 | 4 | 234 | 0 | 252 |
| $y_6$ | contour chipping | 0 | 10 | 325 | 183 | 744 | 27 | 561 | 5 | 94 | 79 | 7 | 2035 |
| $y_7$ | superficial chipping | 0 | 6 | 1 | 3 | 20 | 7 | 43 | 263 | 2 | 3 | 0 | 348 |
| $y_8$ | ambiguity | 2 | 0 | 0 | 2 | 0 | 1 | 0 | 14 | 0 | 3 | 5 | 27 |
| $y_9$ | plane chipping | 4 | 0 | 0 | 2 | 1 | 7 | 2 | 6 | 0 | 0 | 0 | 22 |
| $y_{10}$ | light inking | 0 | 0 | 0 | 0 | 8 | 0 | 26 | 1 | 0 | 338 | 2 | 375 |
| $y_{11}$ | triangular row | 0 | 0 | 0 | 0 | 0 | 0 | 0 | 4 | 0 | 0 | 0 | 4 |
| $y_{12}$ | bump | 0 | 0 | 0 | 0 | 0 | 0 | 0 | 0 | 0 | 26 | 342 | 368 |
| $y_{13}$ | fine cracks | 18 | 0 | 0 | 315 | 2 | 119 | 101 | 146 | 3 | 2 | 3 | 709 |
| $y_{14}$ | impurities | 10 | 0 | 0 | 64 | 8 | 850 | 645 | 73 | 15 | 328 | 141 | 2134 |
| $y_{15}$ | chipping | 0 | 34 | 0 | 0 | 0 | 0 | 0 | 0 | 0 | 0 | 0 | 34 |
| $y_{16}$ | abnormal chamfer | 0 | 6 | 0 | 0 | 0 | 0 | 0 | 0 | 0 | 0 | 0 | 6 |
| $y_{17}$ | crack | 0 | 2 | 0 | 0 | 0 | 0 | 0 | 0 | 0 | 0 | 1 | 3 |
| $y_{18}$ | gas hole | 0 | 0 | 0 | 0 | 0 | 0 | 0 | 0 | 0 | 0 | 1 | 1 |
| $y_{19}$ | stains | 0 | 0 | 0 | 2 | 0 | 0 | 0 | 0 | 0 | 0 | 0 | 2 |
| | Samples | 385 | 3165 | 2464 | 9486 | 6171 | 5796 | 6873 | 5444 | 3040 | 2835 | 2169 | 47828 |
| | images | 252 | 3154 | 2443 | 7008 | 5294 | 5134 | 6293 | 4946 | 2792 | 2542 | 2050 | 47561 |

### 3.2 Evaluation metrics for imbalanced data

Although a classifier should offer a balanced degree of predictive accuracy for both the minority and majority classes on the dataset, they usually provide a severely imbalanced degree of accuracy. The reason is that traditional metrics such as precision and recall are focused on the positive category only, avoiding the problems encountered by multi-class focus metrics in the case of long-tailed distributions [29]. Thus, we decided to use both the F-measure and PR-curve. The first measurement is the weighted harmonic mean of precision (P) and recall (R) of a classifier, taking α=1 (F1 score). The PR curve can complement the previous score because it evaluates changes in distributions, observes variability in performance, and is practical in highly skewed domains. The absence of TN in its equation is functional in imbalanced classes like ours.

### 3.3 Implementation details

We implement the proposed algorithm in Python using a Jupyter Notebook run in Ubuntu 20.04. Our experiments were performed on a PC with an Intel(R) Core(TM) i5-10400F 2.90GHz CPU and an NVIDIA RTX 3060 GPU. Regarding neural network models, we utilized YOLOv5 [30] because it offers model scaling, is easy to implement and modify. Its architecture loads pre-trained weights in COCO from their respective official repositories. This model was trained with a batch size of 11 on a single GPU for 50 epochs. All the source code and pre-trained models of this project are available at https://github.com/ErikValle/RTLAOI-DD.

## 3.5 Domain discrepancy scores

In order to prove that the domain discrepancy score delivers a way to select a suitable dataset for transfer learning, we used each of them as a target dataset and the rest as sources as established in Table 2. Notice that some datasets obtained a score equals to zero, which means that we are using the source domain as a target (EMD = 0) or both datasets share only a class ($\delta$ =0). Recall, the lower the score, the greater the similarity between the two datasets. Following this principle, it can be observed from Table 2 that NJ-101, LC-101 and XH-3 are the more similar to DC-1.

Table 2. Domain discrepancy scores

|  |  | Source datasets | | | | | | | | | | |
|---|---|---|---|---|---|---|---|---|---|---|---|---|
|  |  | **dc-1** | **jy-381-2** | **jy-381-4** | **lc-101** | **lc-201** | **nj-101** | **nj-201** | **xh-1** | **xh-2** | **xh-3** | **xh-4** |
| **Target dataset** | **dc-1** | 0 | 13.2222 | 0.0000 | 3.1342 | 7.1206 | 2.5852 | 8.7600 | 6.2698 | 13.6900 | 6.1870 | 13.4656 |
|  | **jy-381-2** | 14.4008 | 0 | 4.3584 | 6.1309 | 2.2382 | 6.9052 | 9.5422 | 9.2553 | 12.0100 | 11.7779 | 22.5224 |
|  | **jy-381-4** | 0 | 6.0989 | 0 | 7.2453 | 2.3069 | 7.6034 | 7.6264 | 6.7784 | 8.6271 | 9.9443 | 18.0893 |
|  | **lc-101** | 3.7638 | 9.3354 | 6.5446 | 0 | 4.6349 | 0.8726 | 3.3026 | 2.0188 | 6.2050 | 3.4617 | 9.3213 |
|  | **lc-201** | 7.9384 | 3.1054 | 2.2474 | 4.2335 | 0 | 3.9278 | 5.5972 | 5.1799 | 8.0777 | 5.4744 | 13.4756 |
|  | **nj-101** | 3.0947 | 8.6234 | 6.1676 | 0.8576 | 4.2141 | 0 | 3.8647 | 2.6163 | 6.3437 | 4.0969 | 9.7979 |
|  | **nj-201** | 11.1672 | 14.4550 | 7.6569 | 3.5589 | 6.4790 | 4.2496 | 0 | 1.8773 | 1.8912 | 1.1713 | 3.3375 |
|  | **xh-1** | 7.2164 | 12.0095 | 6.0675 | 1.8969 | 5.2294 | 2.5016 | 1.6373 | 0 | 3.5316 | 0.9520 | 5.3757 |
|  | **xh-2** | 14.0635 | 16.4956 | 9.3723 | 5.2274 | 8.3172 | 5.4108 | 1.3925 | 3.3588 | 0 | 1.9608 | 2.9308 |
|  | **xh-3** | 10.8744 | 16.4718 | 9.4725 | 3.4586 | 7.6191 | 4.1345 | 1.2351 | 1.2738 | 2.5667 | 0 | 2.8788 |
|  | **xh-4** | 16.3409 | 22.7409 | 11.9447 | 7.8019 | 12.2151 | 7.8602 | 2.2840 | 4.9038 | 2.4237 | 2.2619 | 0 |

We performed a group of experiments to verify the performance of applying our transfer learning strategy in four target datasets using pre-trained weights from the closest source datasets, as shown in Figure *2*. To interpret the outcomes, we refer to the target domain DC-1 and its experiments transferring weights from a pretrained network using: COCO (black line), the closes domain (LC-101) without subset selection (blue line), the previous dataset but adding the subset selection approach (red line), NJ-101 using the last two methods (purple and green lines, respectively), and XH-1 adjusted. Figure 2 shows how LC-101, which has the highest number of samples among the three datasets with the lowest domain discrepancy scores, obtained the best F1 and PR scores, but its outcomes slightly passed the other two inputs.

Table 3. Experiments on selected target datasets from different source datasets and the COCO dataset. They take the pre-trained weights of the YOLOv5 backbone and randomly initialize the rest of the layers.

|  |  | **Subset Selection** | | **Full dataset** | |
|---|---|---|---|---|---|
| **Target** | **Source** | **F1 score** | **PR score** | **F1 score** | **PR score** |
| DC-1 | NJ-101 | 0.67@0.256 | 0.741 mAP@0.5 | 0.54@0.715 | 0.596 mAP@0.5 |
|  | JY-381-2 | 0.23@0.093 | 0.285 mAP@0.5 | 0.21@0.078 | 0.289 mAP@0.5 |
|  | LC-101 | 0.58@0.657 | 0.754 mAP@0.5 | 0.55@0.759 | 0.752 mAP@0.5 |
|  | XH-1 | 0.63@0.142 | 0.76 mAP@0.5 | 0.57@0.617 | 0.67 mAP@0.5 |
|  | COCO | - | - | 0.24@0.337 | 0.256 mAP@0.5 |
| JY-381-4 | JY-381-2 | 0.69@0.526 | 0.777 mAP@0.5 | 0.68@0.506 | 0.742 mAP@0.5 |
|  | LC-201 | 0.68@0.446 | 0.891 mAP@0.5 | 0.68@0.522 | 0.715 mAP@0.5 |
|  | NJ-201 | 0.69@0.494 | 0.726 mAP@0.5 | 0.69@0.529 | 0.907 mAP@0.5 |
|  | COCO | - | - | 0.68@0.483 | 0.699 mAP@0.5 |
| NJ-101 | XH-1 | 0.59@0.122 | 0.698 mAP@0.5 | 0.5@0.509 | 0.624 mAP@0.5 |
|  | LC-101 | 0.52@0.596 | 0.644 mAP@0.5 | 0.57@0.167 | 0.619 mAP@0.5 |
|  | COCO | - | - | 0.5@0.404 | 0.532 mAP@0.5 |
| XH-4 | LC-201 | 0.38@0.618 | 0.408 mAP@0.5 | 0.41@0.381 | 0.437 mAP@0.5 |
|  | NJ-201 | 0.39@0.231 | 0.423 mAP@0.5 | 0.35@0.533 | 0.453 mAP@0.5 |
|  | XH-3 | 0.39@0.415 | 0.396 mAP@0.5 | 0.37@0.378 | 0.411 mAP@0.5 |
|  | COCO | - | - | 0.34@0.515 | 0.386 mAP@0.5 |

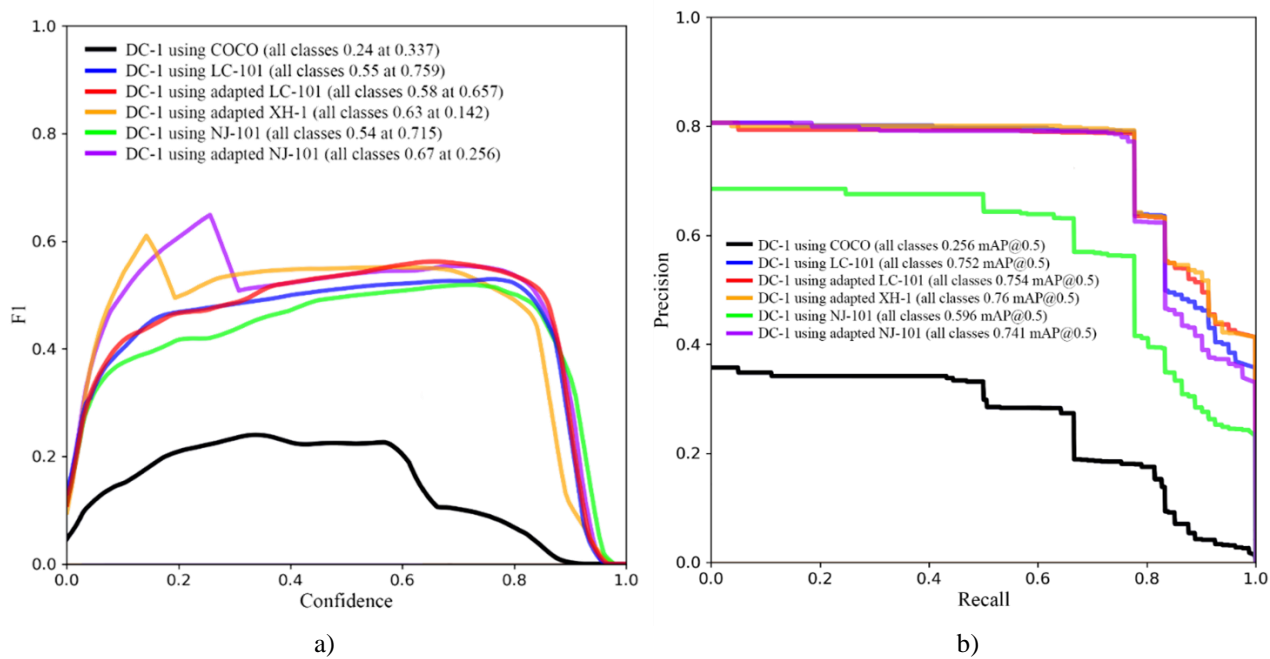

Figure 2 Metrics after transferring pretrained weights using the COCO dataset, LC-101, NJ-101, and our approach. a) F1 scores and b) PR curves.

## CONCLUSIONS

In this study, we proposed a domain discrepancy score to evaluate existing source datasets for a target dataset in terms of the EMD, categories overlapping, the number of samples, and the Gini score. The principal aim was to find highly similar sub-datasets from source datasets to target tasks through our dataset screening based on domain similarity. The network was first pretrained on a benchmark dataset (COCO) and subsequently transferred its backbone weights to the source network, which takes a subset of the source dataset as input. The fine-tuned parameters from the source network belonging to the backbone are assigned to the target network and later fine-tuned on the target dataset. We performed groups of experiments on different magnetic tile datasets to compare our model with typical transfer learning techniques such as full source domain or simply loading pre-trained weight from the COCO dataset. The results show that our method retrieved up to 20% higher F1 and PR scores than using a benchmark dataset as a source domain. In the future, we want to include the number of samples per category in the discrepancy score, add more defect inspection datasets to our experiments, and expand the techniques used in data augmentation to get richer inputs.

## ACKNOWLEDGMENT

This work was supported in part by the Interdisciplinary Foundation of Shenzhen International Graduate School of Tsinghua University (Grant No. JC2021003), in part by the Shenzhen Stable Supporting Program (Grant No. WDZC20200820200655001).

# REFERENCES

[1] A. A. R. M. A. Ebayyeh and A. Mousavi, "A Review and Analysis of Automatic Optical Inspection and Quality Monitoring Methods in Electronics Industry," *IEEE Access,* vol. 8, pp. 183192-183271, 2020.

[2] N. Y. Hammerla, S. Halloran and T. Plötz, "Deep, convolutional, and recurrent models for human activity recognition using wearables," in *25th International Joint Conference on Artificial Intelligence*, New York, 2016.

[3] R. Huang, J. Gu, X. Sun, Y. Hou and S. Uddin, "A Rapid Recognition Method for Electronic Components Based on the Improved YOLO-V3 Network," *Electronics,* vol. 8, no. 8:25, 2019.

[4] Y. Huang, C. Qiu and K. Yuan, "Surface defect saliency of magnetic tile," *The Visual Computer,* vol. 36, pp. 85-96, 2020.

[5] N. Anwar, Z. Shen, Q. Wei, G. Xiong, P. Ye, Z. Li, Y. Lv and H. Zhao, "YOLOv4 Based Deep Learning Algorithm for Defects Detection and Classification of Rail Surfaces," *2021 IEEE International Intelligent Transportation Systems Conference (ITSC),* pp. 1616-1620, 2021.

[6] J. Xing and M. Jia, "A convolutional neural network-based method for workpiece surface defect detection," *Measurement,* vol. 176, pp. 1-15, 2021.

[7] J. Li, Z. Su, J. Geng and Y. Xin, "Real-time Detection of Steel Strip Surface Defects Based on Improved YOLO Detection Network," *IFAC-PapersOnLine,* vol. 51, no. 21, pp. 76-81, 2018.

[8] Y. He, K. Song, Q. Meng and Y. Yan, "An end-to-end steel surface defect detection approach via fusing multiple hierarchical features," *IEEE Transactions on Instrumentation and Measurement,* vol. 69, no. 4, pp. 1493-1504, 2020.

[9] Z. Liu, Z. Miao, X. Zhan, J. Wang, B. Gong and S. X. Yu, "Large-scale long-tailed recognition in an open world," *2019 IEEE/CVF Conference on Computer Vision and Pattern Recognition (CVPR),* pp. 2532-2541, 2019.

[10] H. Lin, B. Li, X. Wang, Y. Shu and S. Niu, "Automated defect inspection of LED chip using deep convolutional neural network," *Journal of Intelligent Manufacturing,* vol. 30, p. 2525–2534, 2019.

[11] J. Tao, Y. Zhu, W. Liu, F. Jiang and H. Liu, "Smooth Surface Defect Detection by Deep Learning Based on Wrapped Phase Map," *IEEE Sensors Journal,* vol. 21, no. 14, pp. 16236-16244, 2021.

[12] G. Fu, P. Sun, W. Zhu, J. Yang, Y. Cao, M. Y. Yang and Y. Cao, "A deep-learning-based approach for fast and robust steel surface defects classification," *Optics and Lasers in Engineering,* vol. 121, pp. 397-405, 2019.

[13] W.-P. Tang, S.-T. Liong, C.-C. Chen, M.-H. Tsai, P.-C. Hsieh, Y.-T. Tsai, S.-H. Chen and K.-C. Wang, "Design of Multi-Receptive Field Fusion-Based Network for Surface Defect Inspection on Hot-Rolled Steel Strip Using Lightweight Dataset," *Applied Sciences,* vol. 11, no. 20, pp. 2076-3417, 2021.

[14] X. He and X. Qian, "A real-time surface defect detection system for industrial products with long-tailed distribution," in *2021 IEEE Industrial Electronics and Applications Conference (IEACon)*, 2021.

[15] S. Kim, W. Kim, Y.-K. Noh and F. C. Park, "Transfer learning for automated optical inspection," in *2017 International Joint Conference on Neural Networks (IJCNN)*, 2017.

[16] T. Han, L. Zhang and S. Jia, "Bin similarity-based domain adaptation for fine-grained image classification," *International Journal of Intelligent Systems,* vol. 37, pp. 2319-2334, 2022.

[17] Q. Yang, Y. Zhang, W. Dai and S. J. Pan, Transfer Learning, Cambridge: Cambridge University Press, 2020.

[18] Y. Rubner, C. Tomasi and L. J. Guibas, "The Earth Mover's Distance as a Metric for Image Retrieval," *International Journal of Computer Vision,* vol. 40, no. 2, pp. 99-121, 2000.

[19] C. Elkan, "Using the Triangle Inequality to Accelerate k-Means," in *Twentieth International Conference on Machine Learning (ICML-2003)*, Washington DC, 2003.

[20] E. J. Russell, "Extension of Dantzig's Algorithm to Finding an Initial Near-Optimal Basis for the Transportation Problem," *Operations Research,* vol. 17, no. 1, pp. 187-191, 1969.

[21] D. Kraft, "A software package for sequential quadratic programming," *Tech. Rep. DFVLR-FB,* vol. 88, no. 28, pp. 30-33, 1988.

[22] C. Gini, "Variabilità e mutabilità," *Journal of the Royal Statistical Society,* vol. 76, no. 3, pp. 326-27, 1913.

[23] L. Yang, H. Jiang, Q. Song and J. Guo, "A Survey on Long-tailed Visual Recognition," in *International Journal of Computer Vision*, 2022.

[24] A. Farahani, S. Voghoei, K. Rasheed and H. R. Arabnia, "A Brief Review of Domain Adaptation," in *Transactions on Computational Science and Computational Intelligence. Advances in Data Science and Information Engineering. Proceedings from ICDATA 2020 and IKE 2020*, Springer, 2021, pp. 877-894.

[25] P. Paraneda Busto and J. Gall, "Open set domain adaption," in *Proceedings of the IEEE International Conference on Computer Vision*, 2017.

[26] H. Liu, Z. Cao, M. Long, J. Wang and Q. Yang, "Separate to Adapt: Open Set Domain Adaptation via Progressive Separation," in *2019 IEEE/CVF Conference on Computer Vision and Pattern Recognition (CVPR)*, 2019.

[27] B. Yang, Z. Liu, G. Duan and J. Tan, "Mask2Defect: A Prior Knowledge-Based Data Augmentation Method for Metal Surface Defect Inspection," *IEEE Transactions on Industrial Informatics,* vol. 18, no. 10, pp. 6743-6755, 2022.

[28] Deutsche Arbeitsgemeinschaft für Mustererkennung, "DAGM 2007," Deutsche Arbeitsgemeinschaft für Mustererkennung, 2007. [Online]. Available: https://conferences.mpi-inf.mpg.de/dagm/2007/index.html. [Accessed 9 September 2022].

[29] H. He and Y. Ma, Imbalanced Learning: Foundations, Algorithms, and Applications, Hoboken, New Jersey: Wiley-IEEE Press, 2013.

[30] G. Jocher, "YOLOv5 Documentation," Ultralytics, 18 May 2020. [Online]. Available: https://docs.ultralytics.com/. [Accessed 17 October 2022].

[31] S. Tang, F. He, X. Huang and J. Yang, "Online PCB defect detector on a new PCB defect dataset," *arXiv:1902.06197v1 [cs.CV],* 2019.

[32] W. Huang and P. Wei, "HRIPCB: a challenging dataset for PCB defects detection and classification," *The 3rd Asian Conference on Artificial Intelligence Technology (ACAIT 2019),* vol. 2020, no. 13, pp. 303-309, 2019.

[33] J. Li, Z. Su, J. Geng and Y. Xin, "Real-time Detection of Steel Strip Surface Defects Based on Improved YOLO Detection Network," *IFAC-PapersOnLine,* vol. 51, no. 21, pp. 76-81, 2018.

[34] Y. Yang, Y. Lou, M. Gao and G. Ma, "An automatic aperture detection system for LED cup based on machine vision," in *Multimedia Tools and Applications*, 2018.

[35] L. Song, W. Lin, Y.-G. Yang, X. Zhu, Q. Guo and J. Xi, "Weak Micro-Scratch Detection Based on Deep Convolutional Neural Network," *IEEE Access,* vol. 7, pp. 27547-27554, 2019.

[36] J. Zhang, S. Li, Y. Yan, Z. Ni and H. Ni, "Surface Defect Classification of Steel Strip with Few Samples Based on Dual-Stream Neural Network," *Steel Research International,* p. 2100554, 2021.

[37] D. He, K. Xu and P. Zhou, "Defect detection of hot rolled steels with a new object detection framework called classification priority network," *Computers & Industrial Engineering,* vol. 128, pp. 290-297, 2019.

[38] Y. Liu, K. Xu and J. Xu, "Periodic Surface Defect Detection in Steel Plates Based on Deep Learning," *Applied Sciences,* vol. 9, no. 15, 2019.

[39] Y.-F. Chen, F.-S. Yang, E. Su and C.-C. Ho, "Automatic Defect Detection System Based on Deep Convolutional Neural Networks," in *2019 International Conference on Engineering, Science, and Industrial Applications (ICESI)*, 2019.

[40] S. T. Rachev, "The Monge–Kantorovich Mass Transference Problem and Its Stochastic Applications," *Theory of Probability and Its Applications,* vol. 29, no. 4, pp. 647-676, 1985.

[41] G. B. Dantzig, Linear Programming and Extensions, Princeton: Princeton University Press, 1963.

[42] H. S. Houthakker, "On the Numerical Solution of the Transportation Problem," *Journal of the Operations Research Society of America,* vol. 3, no. 2, pp. 210-214, 1955.

[43] N. V. Reinfeld and W. R. Vogel, Mathematical Programming, Englewood Cliffs: Prentice-Hall, 1958.

[44] C.-Y. Wang, A. Bochkovskiy and H.-Y. M. Liao, "YOLOv7: Trainable bag-of-freebies sets new state-of-the-art for real-time object detectors," *arXiv:2207.02696 [cs.CV],* 2022.